\documentclass[runningheads]{llncs}
\usepackage{graphicx}
\usepackage[pagebackref=true,breaklinks=true,colorlinks,bookmarks=false]{hyperref}
\usepackage{tikz}
\usepackage{comment}
\usepackage{amsmath,amssymb} %
\usepackage{color}

\usepackage[accsupp]{axessibility}  %

\usepackage[a4paper,scale=0.8]{geometry}

\usepackage{tabulary}
\newcolumntype{x}[1]{>{\centering\arraybackslash}p{#1pt}}
\newcommand{\tablestyle}[2]{\setlength{\tabcolsep}{#1}\renewcommand{\arraystretch}{#2}\centering\footnotesize}
\newlength\savewidth\newcommand\shline{\noalign{\global\savewidth\arrayrulewidth
		\global\arrayrulewidth 1pt}\hline\noalign{\global\arrayrulewidth\savewidth}}
\usepackage{multirow}

\usepackage{pifont}
\usepackage{bm}

\usepackage{xcolor}
\usepackage{graphicx}
\usepackage{amsmath}
\usepackage{amssymb}
\usepackage{booktabs}
\usepackage{subcaption}
\usepackage{floatrow}
\newfloatcommand{capbtabbox}{table}[][\FBwidth]

\usepackage{xspace}
\usepackage{wrapfig}

\makeatletter
\DeclareRobustCommand\onedot{\futurelet\@let@token\@onedot}
\def\@onedot{\ifx\@let@token.\else.\null\fi\xspace}

\def\eg{\emph{e.g}\onedot} 
\def\ie{\emph{i.e}\onedot} 
 
 \def\vs{\emph{vs}\onedot}
 
\def\etal{\emph{et al}\onedot}
\makeatother

\begin{document}
\pagestyle{headings}
\mainmatter

\title{Accelerating the Training of Video \\Super-Resolution Models} %

\titlerunning{ }
\authorrunning{ }

\author{Lijian Lin, Xintao Wang, Zhongang Qi, Ying Shan}
\institute{ARC Lab, Tencent PCG\\ ljlin@stu.xmu.edu.cn, xintao.alpha@gmail.com, \{zhongangqi, yingsshan\}@tencent.com}

\maketitle

\begin{abstract}
Despite that convolution neural networks (CNN) have recently demonstrated high-quality reconstruction for video super-resolution (VSR), efficiently training competitive VSR models remains a challenging problem. 
It usually takes an order of magnitude more time than training their counterpart image models, leading to long research cycles. 
Existing VSR methods typically train models with fixed spatial and temporal sizes from beginning to end. 
The fixed sizes are usually set to large values for good performance, resulting to slow training.
However, is such a rigid training strategy necessary for VSR? 
In this work, we show that it is possible to gradually train video models from small to large spatial/temporal sizes, \ie, in an easy-to-hard manner.
In particular, the whole training is divided into several stages and the earlier stage has smaller training spatial shape. Inside each stage, the temporal size also varies from short to long while the spatial size remains unchanged.
Training is accelerated by such a multigrid training strategy, as most of computation is performed on smaller spatial and shorter temporal shapes. 
For further acceleration with GPU parallelization, we also investigate the large minibatch training without the loss in accuracy. 
Extensive experiments demonstrate that our method is capable of largely speeding up training (up to  $6.2\times$ speedup in wall-clock training time) without performance drop for various VSR models. 
The code is available at \url{https://github.com/TencentARC/Efficient-VSR-Training}.

\end{abstract}

\begin{figure}[t]
	\centering
	\includegraphics[width=0.6\linewidth]{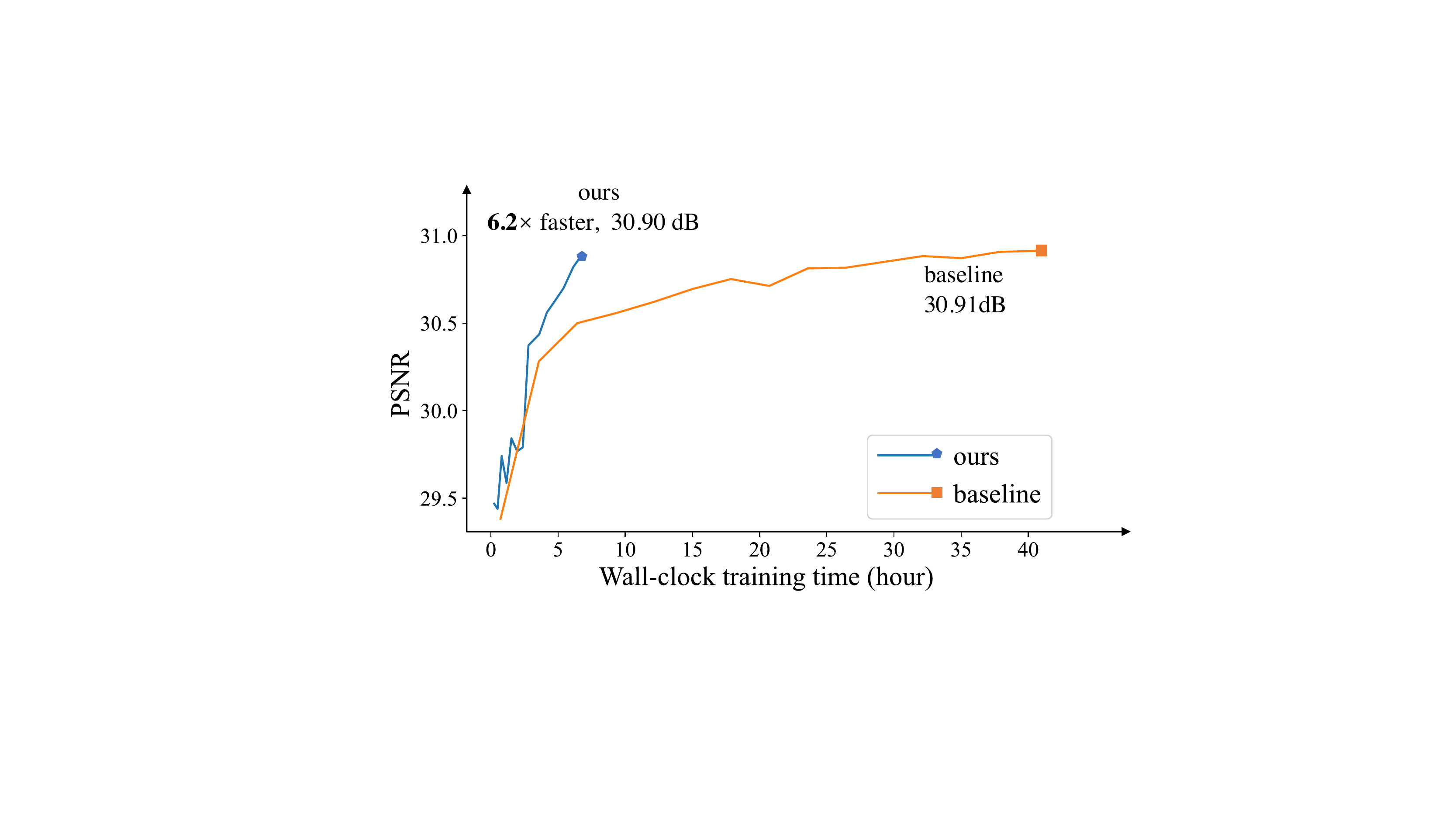}
	\caption{{\bf Wall-clock training time speedup and performance comparisons} on REDS4 with the BasicVSR-M model. Our method significantly accelerates training ({\it i.e.,} $6.2\times$) while maintaining baseline accuracy ($30.90$ \vs $30.91$).  }
	\label{fig:teaser}
\end{figure}

\section{Introduction}
Video super resolution (VSR)~\cite{VSR_1,survey,VSR_9,VSR_10,wang2019edvr,TDAN,VSR_7} aims to recover a high-resolution (HR) video from a low-resolution (LR) input, which has gained increasing attention in computer vision community.
However, training VSR models is much slower than training image SR models due to the additional temporal dimension. The slow training leads to long research cycles, which impedes the development of VSR models.

Existing VSR models~\cite{wang2019edvr,VSR_2,RSDN,chan2021basicvsr} are typically trained with fixed spatial and temporal sizes. The sizes are usually set to large values to achieve good performance.
Larger sizes require the models to process more spatial and temporal information, which is time-consuming. Whereas, training on small sizes is relatively easier and faster, but less accurate. 
It is a natural idea to gradually train VSR models from small to large spatial/temporal sizes, \ie, in an easy-to-hard manner. 
Specifically, in the early stage of training, the VSR models can be trained with small spatial and temporal sizes, which are relatively easier to learn. When the models perform well on small sizes, we then gradually enlarge the spatial and temporal shapes, making the models focus on reconstructing finer details.
Such a learning strategy imitates the way we learn new skills, starting from learning the simple tasks, and then gradually learning the complex and challenging ones.  
In such a way, the training time is largely reduced, as most of computation is performed on small spatial and short temporal shapes. 

Directly applying the above easy-to-hard training strategy to VSR models leads to inferior performance. The reasons are two-fold.  Firstly, the spatial and temporal dimensions in videos are highly correlated. 
Changing the spatial size will affect the learning on temporal dimension, and altering temporal size affects that of spatial. 
Thus, simultaneously varying the spatial and temporal sizes from small to large is not optimal.
Secondly, the learning rate in VSR models usually starts at a large value and then gradually decays to a small one. With such a learning rate scheduler, the learning rate is relatively small when the spatial and temporal sizes are switched to large ones, hindering the learning ability of the models.

In this paper, we propose a simple yet effective multigrid training strategy that learns to reconstruct in an easy-to-hard manner.
The strategy adopts a hierarchical design for altering the spatial and temporal sizes with two cycles. 
Specifically,  we first employ a spatial cycle that varies the spatial size from small to large. 
Then, inside each spatial stage with fixed spatial size, we further employ a temporal cycle that moves the temporal size from short to long. 
In order to fit the different degrees of task difficulty when switching spatial and temporal sizes, we introduce a dynamic learning rate scheduler, where the learning rate is re-started with a large value when the spatial or temporal size changes. 
Large learning rate enhances the exploration ability of the VSR models in transferring from easy tasks (small spatial and short temporal sizes) to harder ones (large spatial and long temporal sizes).  
Experiments demonstrate that our multigrid training strategy in this easy-to-hard manner achieves significant speedup in wall-clock training time without losing accuracy.

In order to further accelerate the training of VSR, we resort to making full use of the GPU parallelism by large minibatch training. 
It has been widely explored in high-level vision tasks to accelerate training without accuracy loss~\cite{weird_parallelizing,train_imagenet_1hour,syncSGD}. However, large minibatch training is still under investigation in VSR.
In this paper, we revisit the training of VSR and study how larger minibatch sizes affect the training of VSR. 
Similar to~\cite{train_imagenet_1hour}, we apply a linear scaling rule to adjust the learning rate according to minibatch sizes. In addition, it is necessary to have a warmup phase that trains the network with a small learning rate early in training. As a result, each training iteration can process more samples, leading to faster training.

In summary, we make the following contributions:
\begin{itemize}
	\item We propose a multigrid training strategy for efficient VSR training. This strategy trains the VSR models in an easy-to-hard manner by varying the spatial and temporal sizes from small to large.
	
	\item Large minibatch training is investigated in VSR to effectively accelerate the training of VSR models.
	
	\item Extensive experiments on various VSR models demonstrate the effectiveness and generalization of multigrid training and large minibatch training. Especially, our method is capable of achieving up to $6.2\times$ speedup in wall-clock training time while maintaining accuracy for recently VSR models. 
\end{itemize}
\vspace{-2pt}

\section{Related Work}
\noindent
{\bf Video Super Resolution.}
Recently, convolutional neural networks (CNN) have brought great improvements in both image \cite{SISR_1,SISR_2,SISR_3,SISR_4,SRGAN,GFPGAN,SISR_6} and video super-resolution \cite{VSR_1,VSR_2,VSR_3,VSR_4,VSR_5,VRT,basicvsr++,understandingDCN}. Existing VSR methods can be roughly classified into two types: sliding-window-based methods \cite{wang2019edvr,sliding_1,sliding_2,sliding_3,TDAN,DUF}, and recurrent-based methods \cite{chan2021basicvsr,RSDN,RRN,train4K}. 
Sliding-window methods tend to restore a single frame using several neighboring frames within a temporal window. Several sliding-window methods~\cite{sliding_1,sliding_2,sliding_3}  adopt optical flow between frames to guide the spatial warping for temporal alignment. Since the optical flow of low-resolution frames is hard to be estimated, TDAN \cite{TDAN} proposes to align different frames in the feature level with deformable convolutions (DCNs) \cite{DCN,DCN_v2}. EDVR \cite{wang2019edvr} further designs a pyramid alignment module to perform alignment in a coarse-to-fine manner and a fusion module to fuse the features of different frames. 

As one of the representative recurrent-based methods, RSDN \cite{RSDN} proposes a structure-detail block and a hidden state adaptation module to exploit previous frames to super-resolve the LR frame. 
BasicVSR \cite{chan2021basicvsr} adopts a bidirectional recurrent design to propagate the information in videos and employs a simple flow-based alignment to align the features, achieving state-of-the-art performance. Despite their promising performance, the long training time hinders the development of VSR models. In this paper, we aim at accelerating the training of VSR models without a performance drop. We evaluate the effectiveness of the proposed training strategy on both the sliding-window-based ({\it i.e., } EDVR) and the recurrent-based ({\it i.e., } BasicVSR) VSR methods.

\noindent
{\bf Curriculum Learning.}
Curriculum learning is a training strategy that trains machine learning models from easy to hard, which imitates the learning order in human curricula. Researchers have exploit its powers in increasing the convergence speed and improving the performance over various tasks, \eg, object detection~\cite{curriculm_det_2015,curriculm_det_2017,curriculm_det_2018} and neural machine translation~\cite{curriculum_survey,curriculm_NLP_2017,curriculm_NLP_2019}.
Among the works in curriculum learning,
Bengio \etal~\cite{curriculum_learning_2009} trains machine learning models by gradually increasing the complexity of training samples.
The work in~\cite{progressiveGAN} proposes to gradually increase the model complexity by adding new layers during training, which both decreases the training time and achieves better performance.
Note that, our easy-to-hard training strategy can be treated as a type of curriculum learning, which has not been investigated in VSR.

\noindent
{\bf Efficient training.}
Recently has witnessed great success in accelerating training in high-level vision tasks ({\it e.g.,} image classification, object detection)~\cite{kaiming_multigrid,train_imagenet_1hour,efficient_1,efficient_2,efficient_3,efficient_4,efficient_5} . The work in \cite{train_imagenet_1hour} presents a linear scaling rule that speeds up training by using large minibatches. 
 Wu \etal \cite{kaiming_multigrid} propose to accelerate the training of video action recognition models with variable minibatch shapes, which achieves a significant speedup in wall-clock training time. The work in \cite{efficient_GCN} designs a layer-wise training framework for graph convolution networks \cite{GCN} that disentangles the feature aggregation and feature transformation during training, leading to a great reduction of time and memory consumption.

Despite the success of the above-mentioned works in accelerating training, how to accelerate the training of VSR models has still barely been investigated. In this paper, we revisit the training of VSR models and present two effective techniques to speed up VSR training while maintaining accuracy.

\section{Method}
In this section, we first present our multigrid training strategy in Sec~\ref{sec:multi_grid}. Then, we introduce how to train VSR models with large minibatches in Sec~\ref{sec:batch_size}.

\begin{wrapfigure}{R}{0.61\linewidth}
	\centering
	\begin{subfigure}{0.50\linewidth}
		\includegraphics[width=\linewidth]{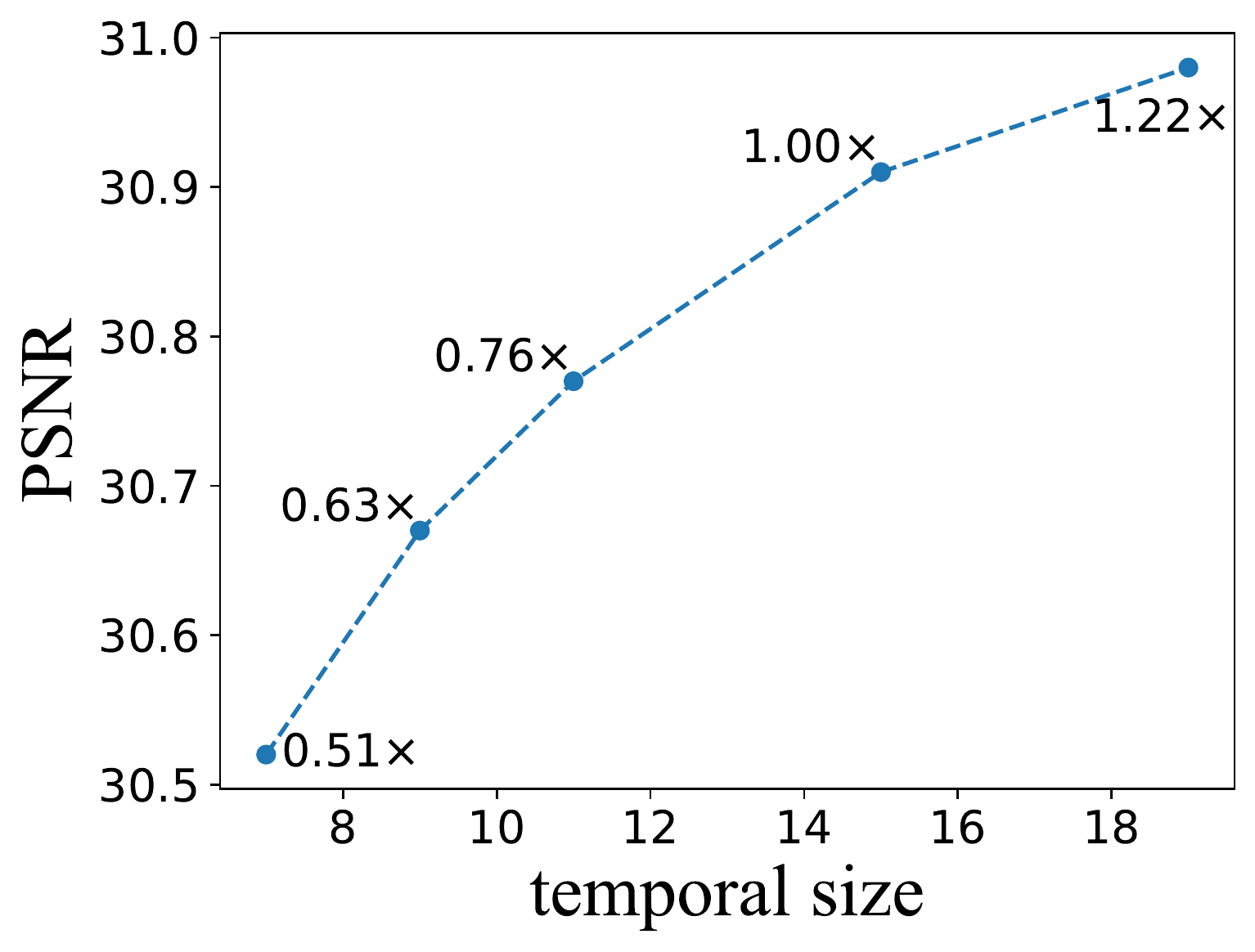}
		\caption{different temporal sizes}
	\end{subfigure}
	\hfill
	\begin{subfigure}{0.48\linewidth}
		\includegraphics[width=\linewidth]{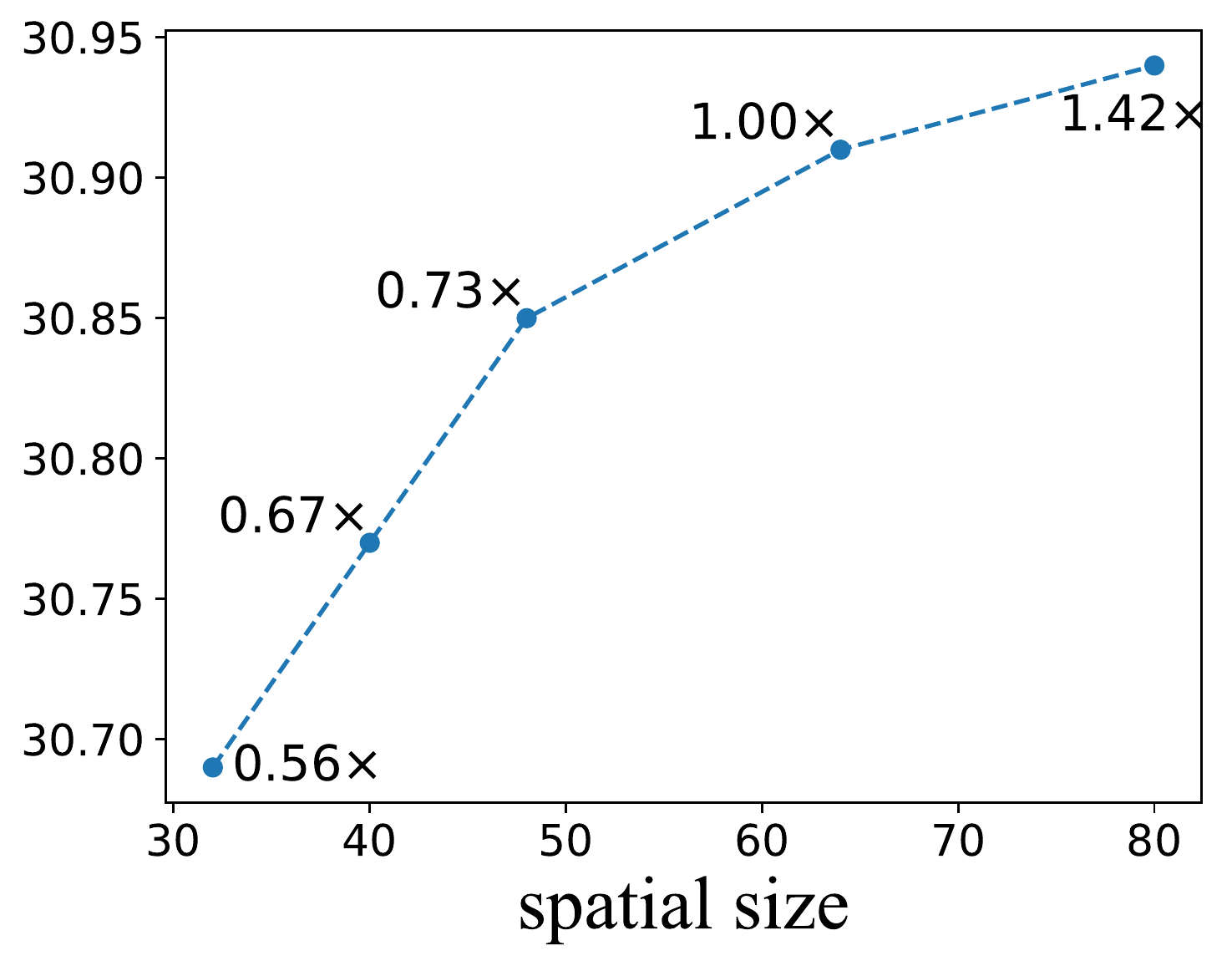}
		\caption{different spatial sizes}
	\end{subfigure}
	\caption{{\bf PSNR performance \vs different temporal size (a),  and different spatial size (b).}  `$\cdot \times$' indicates the relative wall-clock time speedup compared to baseline ($1.00\times$). }
	\label{fig:different_ST}
\end{wrapfigure}
\vspace{-10pt}

\begin{figure*}[h]
	\centering
	\includegraphics[width=\linewidth]{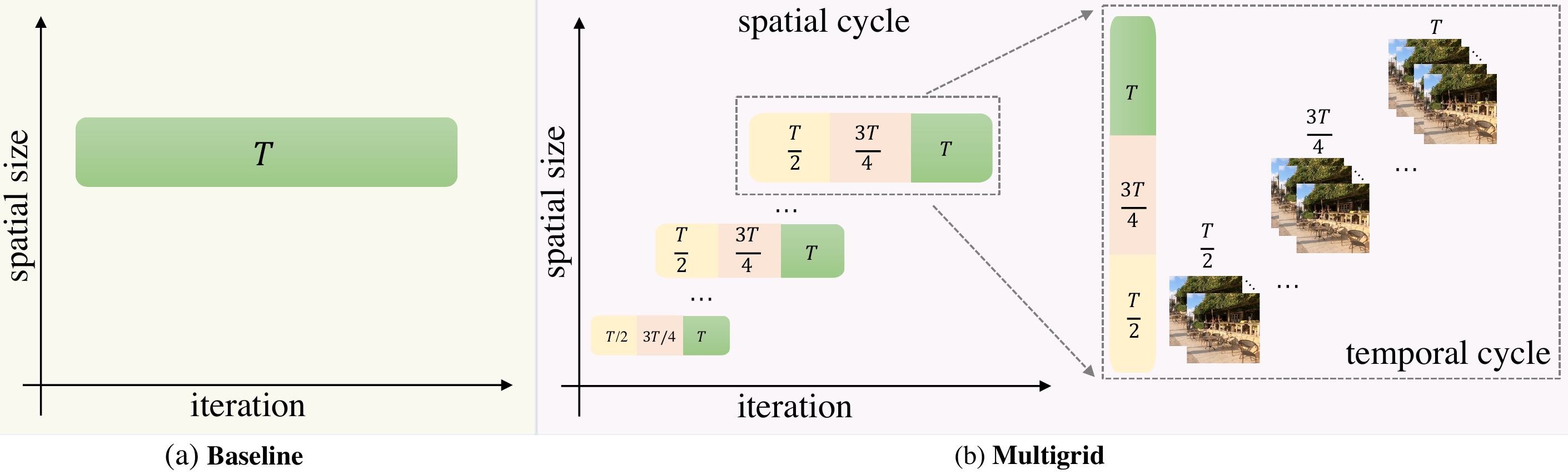}
	\caption{{\bf Training process of the proposed multigrid training strategy \vs baseline training.} (a) {\bf Baseline} training typically adopts a constant and large spatial/temporal size during the whole training process. (b) {\bf Multigrid} training first varies the spatial size from small to large as the training processes. Then, for each stage with fixed spatial size, the temporal size moves from small to large as well. Yellow, pink, and green indicate temporal sizes $\frac{T}{2}$,  $\frac{3T}{4}$, and $T$, respectively.}
	\label{fig:pipeline}
\end{figure*}

\subsection{Multigrid Training}
\label{sec:multi_grid}
Despite the success of image SR methods, directly applying image models to videos leads to inferior performance, as they process each frame separately and thus ignore the rich information among frames.
A common practice to improve the accuracy of VSR is to train the SR methods with multiple frames ({\it i.e.,} large temporal size).  However, as the number of frames grows, training becomes slower (Figure \ref{fig:different_ST}(a)), as the models need to process more temporal information in one forward.
Similarly, larger spatial size yields better VSR performance but long training time (Figure \ref{fig:different_ST}(b)). 
It is natural to raise the question:
{\it is it necessary to keep the spatial and temporal sizes large and fixed during the whole training process?} 
In this paper, we show that the answer is {\it No}.
An intuitive idea is to first train the VSR models at small spatial and temporal sizes, and then gradually switch to larger ones, {\it i.e.,} in an easy-to-hard manner. Specifically, in the early stage of training, the network is trained with small spatial and temporal sizes, which is relatively easier and faster. 
However, it suffers from limited information contained in small sizes, leading to inferior performance. 
One can improve the performance by increasing the spatial and temporal sizes, due to larger sizes make the network focus on fusing more information and reconstructing finer details. 
In such a way, most of the training iterations are conducted with smaller spatial and shorter temporal sizes, leading to faster training.

\noindent
{\bf Multigrid Training Strategy.}
Motivated by the above discussions, we propose a multigrid VSR training strategy that varies the spatial and temporal sizes from small to large throughout training.
Figure \ref{fig:pipeline} illustrates the overview of our multigrid training strategy.
This strategy adopts a hierarchical design with two cycles for altering spatial and temporal sizes.
Specifically, the whole training is divided into several stages and the earlier stage has smaller training spatial shapes. Inside each stage, the temporal size also varies from short to long while spatial size remains unchanged.
Next, we will introduce the details of the proposed spatial cycle and temporal cycle.

\noindent
{\bf Spatial Cycle.}
For the spatial cycle, there exists large design space for 1) different spatial sizes, and 2) duration of each spatial stage. 
Intuitively, the training will be faster if the spatial size starts at smaller values.
However, training with a very small spatial size ({\it e.g.,} $16 \times 16$) leads to a large accuracy drop. The reason might be the unsatisfying optical flow estimation due to the insufficient information produced by such small patches. Therefore, the spatial sizes in the spatial cycle should not be too small. Moreover, in order to achieve baseline accuracy, we set the spatial size in the last spatial stage to the default size ($H \times W$) used in the baseline.
For simplicity, we equally divide the whole training process into $s$ spatial stages, each trained with a fixed spatial size.
The effects and results of different combinations of spatial sizes are provided in Sec~\ref{sec:ab_multi}. 

\noindent
{\bf Temporal Cycle.}
Similarly, the challenge of designing a temporal cycle lies in two aspects: 1) different temporal sizes, and 2) duration of each temporal stage. As larger temporal sizes yield longer training time, a natural desire is to start training with a smaller temporal size. However, the performance of VSR drops a lot with a very small temporal size ({\it e.g.,} $3$), as too few adjacent frames could not provide enough complementary information.
Therefore, we start the temporal cycle with a temporal size not less than $6$ and gradually enlarge it until reaches the original temporal size $T$ in the baseline. 
We equally divide each temporal cycle into $f$ temporal stages. The effects and results of different combinations of spatial sizes are provided in Sec \ref{sec:ab_multi}. 

Moreover, our experiments suggest that directly increasing the spatial and temporal sizes at the same time leads to sub-optimal results. 
Thus, rather than changing them synchronously, 
we adopt a hierarchical design with two cycles for altering spatial and temporal sizes.
In particular, for each spatial stage, the temporal sizes will also be varied through a complete temporal cycle, leading to a total of $p = s \times f$ spatial-temporal stages in the whole training process.

\noindent
{\bf Dynamic Learning Rate Scheduler.}
Simply applying the above multigrid strategy to VSR training causes an accuracy drop. The devil is the learning rate scheduler. Typically, the learning rate in VSR training will be initialized with a relatively large value and then decayed as training progresses. 
If we apply the multigrid training into a baseline VSR network using the original learning rate scheduler, the learning rate in training larger spatial and temporal sizes will be smaller. The small learning rate hinders the exploration ability of the network when spatial and temporal sizes are switched to larger ones. 

In this paper, we propose a dynamic learning rate scheduler, which adjusts the learning rate to fit the different degrees of task difficulty when switching spatial/temporal sizes. 
Specifically, the scheduler re-starts the learning rate with large values when the spatial or temporal size changes.
Following previous practice~\cite{wang2019edvr,chan2021basicvsr}, we apply the cosine annealing strategy for better convergence. In the multigrid training,  the learning rate $\eta_t$  at iteration $t$ is formulated as follows:
\begin{equation}
	\eta_t =
	\begin{cases}		
		cos(\frac{t - \sum_{j=1}^{s(t)-1}P_j}{I_{total}}) \times \eta, \ \ \  0 < s(t) \leq p-1, \\ \\
		cos(\frac{t - \sum_{j=1}^{p-1}P_j}{P_p}) \times \eta,\ \ \ \      s(t) = p-1
	\end{cases},
\end{equation}
where, $\eta$ indicates the initial learning rate used in baseline. 
$P_j$ represents the number of training iterations for spatial-temporal stage $j$, and $I_{total}$ is the total training iterations, satisfying: $\sum_{j=1}^p P_j = I_{total}$.  
$s(t) \in \{1, 2, ..., p\}$ indicates that the iteration $t$ belongs to the $s(t)$ spatial-temporal stage, 
\ie, when $0\leq t < P_1, s(t) = 1$.
Since the total iterations $I_{total}$ is always larger than $t- \sum_{j}^{s(t)-1}P_j$, the learning rate $\eta_t$ will never decay to zero for ${P_1, P_2, ..., P_{p-1}} $, which avoids wasting training iterations with too small learning rates.

Wu {\it et al.} \cite{kaiming_multigrid} propose to efficiently train video recognition networks by making the minibatch shape (minibatch size, spatial size, and temporal size) variable. They change the minibatch shapes following a fundamental concept that the magnitude of information at one iteration needs to be unchanged, which does not follow the easy-to-hard learning protocol. Thus, they use the default learning rate scheduler.  Different from their practice, we aim to accelerate the training of VSR networks by varying the spatial and temporal sizes from small to large to perform easy-to-hard learning. That means the magnitude of information differs a lot in different spatial-temporal stages. Therefore, rather than using the default learning rate scheduler, we design a proper learning rate scheduler for our efficient VSR training.

\subsection{Large Minibatch Training}
\label{sec:batch_size}

Recently, researchers have made significant developments in accelerating training by increasing minibatch sizes in high-level vision tasks \cite{train_imagenet_1hour,weird_parallelizing} ({\it e.g.} image classification, object detection). Increasing minibatch sizes enables a network to process more samples in parallel, thus they can train the same number of epochs faster. However, how to train VSR networks faster by using larger minibatch sizes while maintaining accuracy has barely been investigated. In this paper, we investigate the large minibatch training for VSR. 
Similar to the practice developed in high-level tasks~\cite{train_imagenet_1hour}, we conclude two important rules for accelerating training with large minibatches. 1) Linearly scale the learning rate when the minibatch size changes. 2) Warmup the network with a smaller learning rate at the beginning.

Next, we will review the training of VSR networks to discuss why the above-mentioned rules are effective. We consider a typical VSR training with minibatch using  the loss $L(w)$ : 
\begin{equation}
	L(w) = \frac{1}{n} \sum_{x \in X} l(x, w), 
	\label{loss}
\end{equation}
where $X$ is a minibatch and $n = |X|$ indicates the number of samples in $X$ ({\it i.e., } minibatch size).  $w$ is the weights of a VSR network. $l(\cdot, \cdot)$ is the loss between the output of the network and ground truth. 

We analyze the differences between training $m$ iterations with $m$ small minibatches $X_{0-m}$, and training a single iteration with one large minibatch $\bm{\mathcal{X}}$.  Each of the minibatch $X_{0-m}$ holds $n$ samples. $\bm{\mathcal{X}}$ consists of those small minibatches $X_{0-m}$, which means $|\bm{\mathcal{X}}| = mn$. According to Eq.~\ref{loss},  after $m$ iteration of training using $m$ minibatches $X_{0-m}$, the weights are updated as follows:
\begin{equation}
	w_{t+m} = w_t - \eta \frac{1}{n} \sum_{i}^{m} \sum_{x \in X_i} \nabla l(x, w_{t+i}), 
\end{equation}
where $\eta$ is the learning rate and $t$ indicates the training iteration index.  $\nabla l$ is the gradient according to loss $l(\cdot, \cdot)$. Similarly, when executing a single iteration with minibatch $\bm{\mathcal{X}}$ of size $mn$, the weights will be:
\begin{equation}
	w_{t+1}' = w_t - \eta'\frac{1}{mn} \sum_{x \in \bm{\mathcal{X}}} \nabla l(x, w_t). 
\end{equation}
Note that if we assume for $i < m$, $w_i \approx w_{i+m}$, then:
\begin{equation}
	\sum_{i}^{m} \sum_{x \in X_i} \nabla l(x, w_{t+i}) \approx \sum_{x \in \bm{\mathcal{X}}} \nabla l(x, w_t).
\end{equation}
Thus, when the learning rate is set as $\eta' = m\eta$, we will have $w_{t+m} \approx w_{t+1}'$. This indicates that we can train the network with larger minibatch sizes and fewer iterations to approximate the baseline training with linear scaled learning rate. 

Besides, this linear scaling rule relies on the assumption that $w_i \approx w_{i+m}$. This assumption might fail when the weights change rapidly. Since the rapid changing of weights usually occurs in the early stages of training, we apply a warmup strategy that gradually increases the learning rate from a small value to a large one to alleviate this issue.

\section{Experiments}

\subsection{Implementation Details}

\noindent
{\bf Spatial Cycle.}
We equally divide the training process into $s=2$ spatial stages. The spatial sizes in these two stages are set to $max(32 \times 32, \frac{H}{2} \times \frac{W}{2})$ and $H \times W$, respectively. Training samples with different spatial sizes are generated by randomly cropping the original frames.

\noindent
{\bf Temporal Cycle.} For each spatial stage in the spatial cycle, we further equally divide it into $f=3$ temporal stage. The temporal sizes ({\it i.e.,} number of consecutive frames fed into VSR models) in these three temporal stages are set in an increasing way: $max(6, \frac{T}{2})$, $\frac{3T}{4}$ and $T$. These three temporal sizes cover an intuitive range and work well in practice. By doing so, there will be $p = 2 \times 3 = 6$ spatial-temporal stages in total during the whole training process. 

\noindent
{\bf Learning Rate Scheduler.}
Our dynamic learning rate scheduler consists of $p$ periods, which are synchronized with the above-mentioned $p$ spatial-temporal stages. For each period, the learning rate begins at a large value (the initial learning rate used in baseline) and then decays following the cosine annealing~\cite{cosine}. 

\noindent
{\bf Datasets and Evaluation Metrics.}
We conduct our experiments on the REDS \cite{REDS} and Vimeo-90K \cite{vimeo-90k} datasets, which are widely-used and challenging datasets for VSR. REDS contains $300$ video clips with a total of $300, 000$ frames. Following \cite{chan2021basicvsr,wang2019edvr}, we adopt REDS4 as our test set, and use the left as training set.
Vimeo-90K contains $64,612$ training, and $7,824$ testing $7$-frame video sequences.
All the datasets are commonly used in VSR and licensed for research purposes.
The performance is measured in terms of PSNR and SSIM. As we use remote data access, the unstable data loading highly affects the measure of wall-clock training time. Thus, we report the wall-clock training time without the data loading time.

\noindent
{\bf Training and Inference Details.}
We adopt two VSR models to evaluate the effectiveness of the proposed multigrid training and large minibatch training: BasicVSR-M \cite{chan2021basicvsr}, BasicVSR \cite{chan2021basicvsr}, and EDVR-M \cite{wang2019edvr} (M denotes the medium size). For BasicVSR-M and BasicVSR, the spatial sizes used in the spatial cycle are: $32 \times 32$ and $64 \times 64$ on both REDS and Vimeo-90K. The temporal sizes in the temporal cycle are \{$7, 11, 15$\} and \{$6, 10, 14$\} on REDS and Vimeo-90K, respectively. Note that EDVR-M adopts a sliding window design, where the temporal size is determined by its model architecture, and usually cannot be changed for both training and testing. Thus, we only apply the spatial cycle to it. 
The spatial sizes for EDVR-M are $32 \times 32$ and $64 \times 64$ on both REDS and Vimeo-90K. 
The learning rate of training on $32 \times 32$ spatial size begins at $2e-4$, as too large learning rate may cause severe performance drop for EDVR-M.
In addition, we train these models with linear learning rate warmup for the first $5,000$ iterations. The training and analyses are performed with PyTorch on NVIDIA V100 GPUs in an internal cluster.

\subsection{Experiments on REDS}
\label{sec:reds}

\begin{table*}[t]
	\centering
	\caption{{\bf Quantitative comparison on REDS4 with BasicVSR.} We report the wall-clock speedup relative to baseline (\ding{173}-\ding{176} \vs \ding{172}, and \ding{178}-\ding{181} \vs \ding{177}, respectively).  * means the results are collect from the original paper.
	Best performance is highlighted with \textbf{bold}. `S' and `T' denote spatial and temporal, respectively.
	\label{tab:main_basicvsr}
}	
		\tablestyle{6.5pt}{1.2}
	\begin{tabular}{l|cccc|cc}
		model & id & minibatch & multi & speedup & PSNR & SSIM \\
		\shline
		BasicVSR-M &\ding{172} & 16 & - & - & 30.91 & 0.8824 \\  
		Large-Batch & \ding{173} & 64 & - & 3.9$\times$ &  30.93 & 0.8824   \\
		\hline
		Multi-S & \ding{174} & 64 & S & 4.9$\times$ & 30.90 & 0.8822 \\
		Multi-T & \ding{175} & 64 & T & 5.0$\times$ & 30.93 & 0.8830    \\
		{\bf ours}& \ding{176} & 64 & S\&T & \textbf{6.2$\times$} &  30.90  & 0.8820 \\
		\shline
		BasicVSR* &  \ding{176} &32 & - & - & 31.42 & 0.8909 \\
		BasicVSR(our impl.) & \ding{177}& 32 & - & - &  31.58 & 0.8934 \\
		Large-Batch& \ding{178} & 64 & - &  1.9$\times$ &  31.62 & 0.8943  \\
		\hline
		Multi-S & \ding{179} & 64 & S & 2.4$\times$ &  31.56 & 0.8932\\
		Multi-T & \ding{180} & 64 & T & 2.5$\times$ &  31.61 &  0.8941\\
		{\bf ours}& \ding{181} &64 & S\&T & \textbf{3.1$\times$} & 31.54 & 0.8925
	\end{tabular}
	\vspace{-10pt}
\end{table*}

\begin{table*}[t]
		\caption{{\bf Quantitative comparison on REDS4 with EDVR-M.} We report the wall-clock speedup relative to baseline training (\ding{174}-\ding{175} \vs  \ding{173}).  * means the results are collect from the original paper.
		Best performance is highlighted with \textbf{bold}. `S' and `T' denote spatial and temporal, respectively.
		\label{tab:main_edvr}
	}
	\centering
	\tablestyle{7pt}{1.2}
	\begin{tabular}{l|cccc|cc}
		model & id & minibatch & multi & speedup & PSNR & SSIM \\
		\shline
		EDVR-M* &\ding{172}  & 32 & - & - & 30.46   &   0.8684\\
		EDVR-M (our impl.) &\ding{173}  & 32 & - & - &  30.45 &  0.8687  \\
		Large-Batch &\ding{174}  & 64 & - & 1.9$\times$  & 30.46 &  0.8689  \\
		\hline
		{\bf ours} &\ding{175}  & 64 & S & \textbf{2.3$\times$} &  30.44 &  0.8685\\
	\end{tabular}
\vspace{-10pt}
\end{table*}

In this subsection, we compare multigrid training and large minibatch training with baseline training on REDS.

\noindent{\bf Results on BasicVSR-M and BasicVSR.}
The quantitative results obtained by BasicVSR-M and BasicVSR are summarized in Table \ref{tab:main_basicvsr}. Applying multigrid training and large minibatch training to BasicVSR-M and BasicVSR achieves significant speedup ({\it i.e.,} $6.2\times$, $3.1\times$ for BasicVSR-M and BasicVSR, respectively) without losing accuracy. Specifically, 1) the training becomes $3.9\times$ and $1.9\times$ faster when training BasicVSR-M and BasicVSR with $4\times$ and $2\times$ larger minibatches. The speedup can be attributed to that large minibatch sizes enable better GPU parallelization. 
2) The training time can be further reduced by employing the proposed multigrid training strategy (see Table \ref{tab:main_basicvsr} \ding{174}-\ding{176}, \ding{179}-\ding{181}). Table \ref{tab:main_basicvsr} shows that both the spatial cycle and temporal cycle bring consistent speedup to BasicVSR with different model sizes. Moreover, combining them together ({\it i.e.,} the proposed multigrid training strategy) achieves the fastest training while maintaining accuracy. These results suggest that a VSR model can be efficiently trained in an easy-to-hard manner ({\it i.e.,} from small spatial/temporal sizes to larger ones), and finally reach the baseline performance. 

\noindent
{\bf Results on EDVR-M.}
Next, we apply the multigrid training and large minibatch training to a sliding-window-based method EDVR-M to investigate their generalization to different VSR models. As shown in Table \ref{tab:main_edvr}, training EDVR-M with the multigrid training strategy and large minibatch leads to a significant $2.3\times$ speedup. We observe that the speedup is consistent with the recurrent-based method  BasicVSR. These results demonstrate that the two proposed strategies are robust and can be easily generalized to different VSR models.

\noindent
{\bf Qualitative Results.}
We also present some qualitative results obtained by our method and baseline in Figure \ref{fig:main_qualitative}. The methods with our multigrid training and large minibatch training obtain comparable visual results to the baselines.

\begin{figure*}[t]
	\centering
	\includegraphics[width=\linewidth]{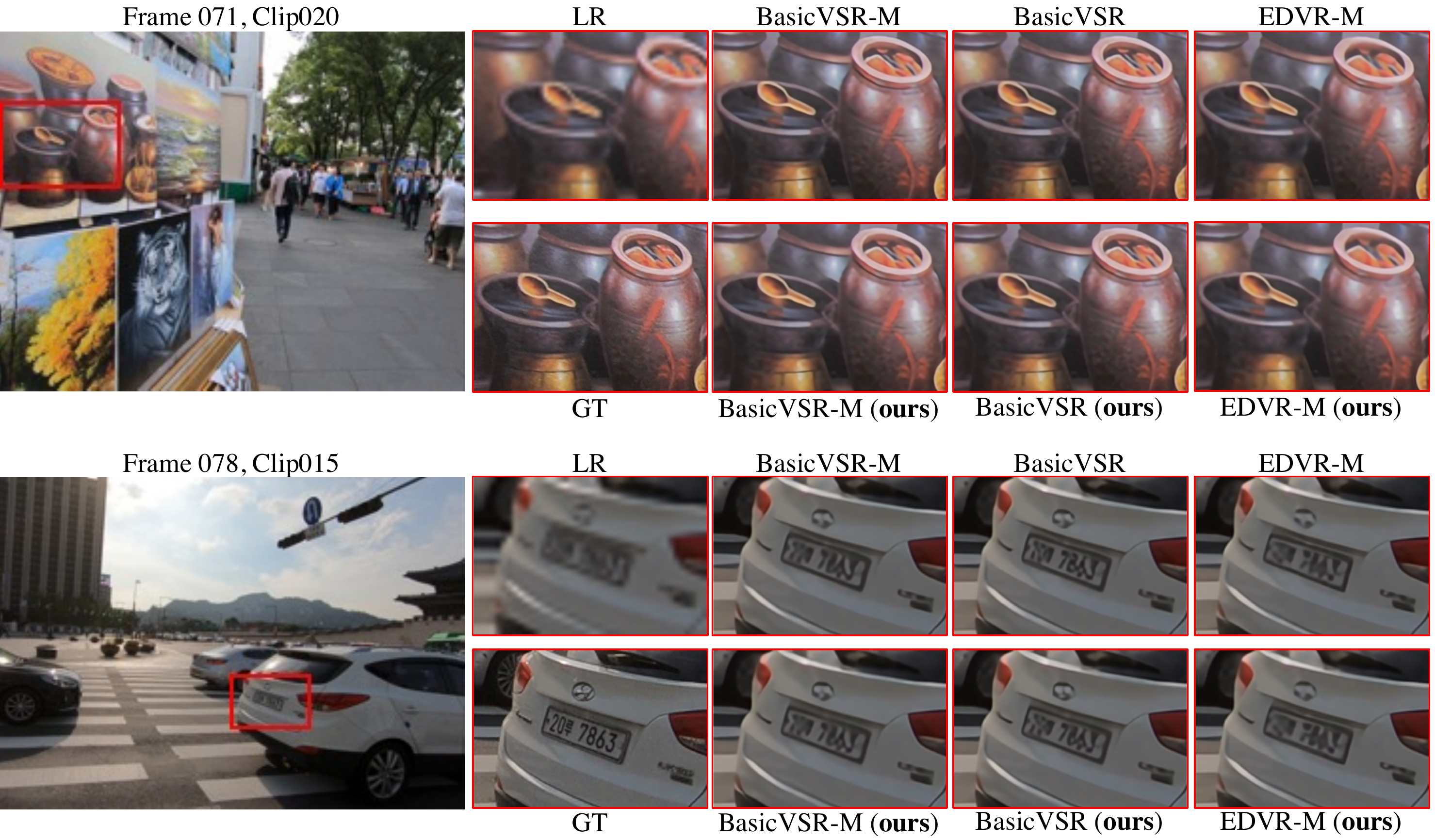}
	\caption{{\bf Qualitative results on REDS4} for 4$\times$ VSR on BasicVSR-M, BasicVSR, and EDVR-M models. The methods with our proposed multigrid training and large minibatch training achieve comparable visual results to the baselines. \textbf{Zoom in for best view.} }
	\label{fig:main_qualitative}
\vspace{-10pt}
\end{figure*}

\begin{table*}[]
	\caption{{\bf Quantitative comparison on Vimeo-90k.} We report the wall-clock speedup relative to baseline training (\ding{173} \vs \ding{172}, and \ding{175} \vs \ding{174}). * means the results are collect from the original paper.
		Best performance is highlighted with \textbf{bold}. 
		\label{tab:vimeo-90k}
	}
	\centering
	\tablestyle{10pt}{1.1}
	\begin{tabular}{l|cccc}
		model & id & speedup & PSNR & SSIM \\
		\shline
		Baseline (BasicVSR*) & \ding{172} & - & 37.18 & 0.9450 \\
		\textbf{ours} & \ding{173} & 3.2$\times$ & 37.32  & 0.9462 \\
		\hline
		Baseline (EDVR-M) &\ding{174}  & - & 35.10 & 0.9426  \\
		\textbf{ours} &\ding{175} & 2.3$\times$ & 35.22 &   0.9437\\

	\end{tabular}
	\vspace{-15pt}
\end{table*}

\subsection{Experiments on Vimeo-90K}
Next, we evaluate the proposed method over BasicVSR and EDVR-M on Vimeo-90K to investigate its generalization toward different VSR datasets.

\noindent
{\bf Results on BasicVSR.}
As in Table~\ref{tab:vimeo-90k} (\ding{173} \vs \ding{172}), applying the proposed multigrid training and large minibatch training to BasicVSR leads to a significant speedup (\ie, $3.2\times$) while maintaining baseline accuracy.

\noindent
{\bf Results on EDVR-M.}
As in Table~\ref{tab:vimeo-90k} ({\ding{175} \vs \ding{174}}), our strategies bring a significant speedup (\ie, $2.3\times$) for EDVR-M without performance drop.

The speedups on the Vimeo-90K dataset are consistent with that on the REDS dataset mentioned in Sec~\ref{sec:reds} for those two VSR methods. 
These results demonstrate that the proposed multigrid training and large minibatch training are robust and can be easily generalized to both different VSR methods and different VSR datasets.

\section{Ablation Studies and Analysis}
In this section, we conduct comprehensive analysis experiments to investigate how minibatch size and multigrid training affect the training of VSR models. All the experiment results are obtained by the BasicVSR-M on REDS.

\begin{table}[]
		\caption{{\bf Ablation study of learning rate scaling.} All the results are obtained by BasicVSR-M on REDS4.
		\label{tab:ab_lr}
	}
	\centering
	\tablestyle{3pt}{1.1}
	\begin{tabular}{l|x{32}x{32}x{32}x{32}}
		model & minibatch & lr & iteration & PSNR  \\
		\shline
		baseline & 16 & 2e-4 & 300k & 30.91 \\
		\hline
		w/o lr scaling    & 64 & 2e-4 & 75k & 30.64 \\
		w \ \ \ lr scaling  & 64 & 8e-4 & 75k &  \textbf{30.93}
	\end{tabular}
   \vspace{-10pt}
\end{table}

\subsection{Minibatch size {\textbf \vs} Performance}

Here, we experimentally evaluate the effectiveness of the linear learning rate scaling and warmup used in large minibatch training mentioned in Sec \ref{sec:batch_size}.

\noindent{\bf Learning Rate Scaling.} 
As shown in Tabel \ref{tab:ab_lr}, directly increasing the minibatch size leads to a performance drop. Whereas, when conducting learning rate scaling, large minibatch training (with warmup) achieves comparable performance to baseline. This suggests that, with the linear scaled learning rate,  the total gradient of large minibatch is roughly equal to that of small ones.

\begin{table}[h]
	\begin{floatrow}
		\capbtabbox[\linewidth]{
			\caption{{\bf Ablation study of warmup.} All the results are obtained by BasicVSR-M on REDS4.
				\label{tab:ab_warmup}
			}
			\centering
			\tablestyle{0pt}{1.1}
			\small
			\begin{tabular}{l|x{40}x{40}x{40}x{40}}
				model & minibatch & warmup & PSNR \\
				\shline
				baseline & 16 & - & 30.91 \\
				\hline
				w/o warmup & 64 & -    & 30.81 \\
				\multirow{2}{*}{w\ \ \ \  warmup} & 64 & constant & 30.91 \\
				& 64 & linear & \textbf{30.93}
			\end{tabular}
		}
		
		\capbtabbox[\linewidth]{
			\caption{{\bf Large minibatch size \vs baseline}. Each GPU holds 4 samples in these experiments.
				\label{tab:ab_bs}
			}
			\centering
			\tablestyle{0pt}{1.1}
			\small
			\begin{tabular}{l|x{45}x{45}x{45}x{45}}
				model & minibatch & speedup & PSNR \\
				\shline
				baseline & 16 & - & 30.91 \\
				\hline
				\multirow{3}{*}{variants }  & 32  & 2.0$\times$  & 30.92 \\
				& 48 &  2.9$\times$  &  \textbf{30.95}\\
				& 64  &  \textbf{3.9$\times$} & 30.93
			\end{tabular}
		}
		
	\end{floatrow}
\end{table}

\noindent{\bf Warmup.}
We investigate the effect of warmup settings in Tabel~\ref{tab:ab_warmup}. Directly applying the linear scaling rule without warmup results in inferior performance. This is probably due to that the network changes rapidly in the early stage of training. Thus the approximation between a single step on a large minibatch, and several steps on small minibatches may fail (as discussed in Sec~\ref{sec:batch_size}). As Tabel~\ref{tab:ab_warmup} shows, with the help of the warmup phase, the performance of training with a large minibatch can achieve baseline performance.

Moreover, we train BasicVSR-M with different minibatch sizes to evaluate the robustness of large minibatch training. As shown in Tabel \ref{tab:ab_bs}, increasing the minibatch sizes provides a solid and significant speedup while maintaining baseline accuracy. In addition, we observe that the speedup factors almost match the minibatch scaling factors, thanks to the GPU parallelism.

\begin{table*}[h]
	\centering
	\begin{subtable}[t]{0.49\linewidth}
		\tablestyle{3pt}{1.2}
		\small
			\begin{tabular}{l|cccc}
				model & spatial size   & speedup & PSNR \\
				\shline
				baseline & 64 & -& 30.91 \\
				\hline
				\multirow{3}{*}{Multi-S} & 32/64& 4.9$\times$ &30.90 \\
				& 32/48/64& 5.0$\times$ & 30.89    \\
				& 32/40/48/64 & 5.1$\times$& 30.89  \\
				
			\end{tabular}
			\caption{{\bf spatial cycle}}
		\end{subtable}
		\begin{subtable}[t]{0.49\linewidth}
			\tablestyle{3pt}{1.2}
			\small
				\begin{tabular}{l|cccc}
					model & temporal size &  speedup  & PSNR \\
					\shline
					baseline & 15 & - & 30.91 \\
					\hline
					\multirow{3}{*}{Multi-T} & 7/15 & 5.0$\times$&  30.92 \\
					& 7/11/15 & 5.0$\times$& 30.93    \\
					& 7/9/11/15 & 5.2$\times$ & 30.91  \\
				\end{tabular}
				\caption{{\bf temporal cycle}}
			\end{subtable}
			
			\caption{{\bf Ablation study of spatial and temporal cycles.} We evaluate the impact of different spatial cycle (a), and temporal cycle (b) designs.  The sizes are presented according to their orders in training. For example, `32/64' indicates the spatial size begins at $32\times32$ and then is switched to $64\times 64$. Our learning rate scheduler is used in all settings. We report the wall-clock speedup relative to baseline.
				\label{tab:ab_multi}
			}
		\vspace{-15pt}
		\end{table*}

\subsection{Multigrid Training}
\label{sec:ab_multi}
Next, we analyze how multigrid training affects the training of VSR methods. All the following experiments are conducted with large minibatch training.

\begin{table*}[h]
	\centering
	\begin{subtable}[t]{\linewidth}
		\tablestyle{4pt}{1.2}
		\begin{tabular}{l|x{100}x{30}x{30}x{30}}
			model & spatial \& temporal size   & speedup & PSNR \\
			\shline
			baseline & 64\&15 & -& 30.91 \\
			\hline
			\multirow{2}{*}{Multi-S\&T} & 32\&7 / 64\&15 & 5.8$\times$ &30.81\\
			& 32\&7 / 48\&11 / 64\&15& 6.0$\times$  & 30.83   \\			
		\end{tabular}
		\caption{{\bf synchronous}}
	\end{subtable}
	\begin{subtable}[t]{\linewidth}
		\tablestyle{4pt}{1.2}
		\begin{tabular}{l|x{170}x{30}x{30}x{30}}
			model & spatial \& temporal size &  speedup  & PSNR \\
			\shline
			baseline & 64\&15 & - & 30.91 \\
			\hline
			\multirow{2}{*}{Multi-S\&T}& 32\&7 / 32\&15 / 64\&7 / 64\&15 & 6.3$\times$  & 30.86 \\
			& 32\&7\textbf{/}32\&11\textbf{/}32\&15\textbf{/}64\&7\textbf{/}64\&11\textbf{/}64\&15 &  6.2$\times$ & 30.90    \\
		\end{tabular}
		\caption{{\bf hierarchical}}
	\end{subtable}
	
	\caption{{\bf Ablation study of different combinations of spatial and temporal cycles.} We combine the proposed spatial and temporal cycles in two different ways: (a) synchronous: change spatial and temporal sizes at the same time; (b) hierarchical: place the temporal cycle into each spatial stage in the spatial cycle. The sizes are presented according to their orders in training. For example, `32\&7/64\&15' indicates that the training beings with a spatial size of $32\times32$ and a temporal size of $7$, and then is switched to a spatial size of $64\times 64$ and a temporal size of $15$. The proposed learning rate scheduler is used in all settings. We report the wall-clock speedup relative to baseline.
		\label{tab:ab_multi_combine}
	}
\end{table*}

\noindent
{\bf Spatial Cycle.}
We evaluate the performance of the proposed spatial cycle with different combinations of dynamic spatial sizes in Tabel \ref{tab:ab_multi}(a). All the variances of Multi-S are trained with the baseline temporal size ({\it i.e.,} $15$) and the proposed learning rate scheduler. 
As shown in Table \ref{tab:ab_multi}(a), varying spatial size from small to large always achieves the baseline performance for different spatial size schemes. These results show the robustness of changing spatial size during training. In addition, more spatial sizes yield slightly faster training, while having slightly lower performance. In order to get a better trade-off between high performance and faster training, we adopt the `32/64' scheme as our default setting.

\noindent
{\bf Temporal Cycle.}
We summarize the performance of BasicVSR-M trained with different combinations of temporal sizes in Table \ref{tab:ab_multi}(b). All the variances of Multi-T are trained with the baseline spatial size ({\it i.e.,} $64 \times 64$) and the proposed dynamic learning rate scheduler.  Similar to the spatial cycle, training with different combinations of temporal sizes can always accelerate training while maintaining baseline accuracy. We employ the `$7/11/15$' scheme in our temporal cycle, which is a good trade-off between effectiveness and efficiency.

\noindent
{\bf Mulitigrid.}
Next, we investigate how to combine the above two cycles together to further speed up our training. As shown in Table \ref{tab:ab_multi_combine}(a), simply combining the sizes in spatial and temporal cycles in a synchronous way ({\it i.e.,} change the spatial and temporal size at the same time) causes a performance drop. 
We conjecture that the large magnitude of information change brought by simultaneously varied spatial/temporal sizes might hinder the learning process.
Table~\ref{tab:ab_multi_combine} shows that combining the spatial and temporal cycles in a hierarchical way ({\it i.e.,} the proposed multigrid training strategy which places the temporal cycle into each spatial stage in the spatial cycle) leads to $6.2\times$ speedup without losing accuracy. These results demonstrate the effectiveness of our multigrid design.

\subsection{Limitations and Discussions}
Despite our method brings significant speedup to the recurrent-based VSR models ({\it e.g.,} BaiscVSR), the speedup in the sliding-window-based models ({\it e.g.,} EDVR) is limited. This is because the temporal size of a sliding window-based method is determined by its model architecture, and usually cannot be changed for both training and testing. 
Besides, our method is designed for accelerating the training of CNN-based VSR models.
Recently, researchers have proposed several Transformer-based VSR methods \cite{VRT,zhang2022cross,Video_SR_trans}, which differ with the CNN-based ones in many aspects, \eg, model architectures and training techniques~\cite{battle2021,convnet_2020s}.
How our method affects the training of Transformer-based VSR methods needs to be further investigated, which is left as our future work.

\section{Conclusion}
In this paper, we propose to accelerate the training of VSR methods with multigrid training and large minibatch training. Different from existing VSR methods that are trained with fixed spatial and temporal sizes, our multigrid training varies the spatial and temporal sizes from small to large, \ie, in an easy-to-hard manner. The training is accelerated by such a multigrid training strategy, as most of computation is performed on smaller spatial and shorter temporal shapes.  
Moreover, we investigate the large minibatch training without accuracy loss for further acceleration with GPU parallelization. 
Extensive experiments on different methods demonstrate the effectiveness and generalization of our proposed multigrid training and large minibatch training in VSR.

\clearpage
\bibliographystyle{splncs04}
\bibliography{egbib}

\clearpage
\appendix
\begin{center}
		\Large 
		\textbf{Accelerating the Training of Video Super-Resolution Models\\ Supplementary Material}
		\par
\end{center}
	\vspace{2em}

In this supplementary material, we first present several implementation details in Sec~\ref{sec: implement_details}.
We also show the results of multigrid training with additional larger spatial-temporal stage in Sec~\ref{sec:addi_ST}. 
Finally, we provide more qualitative results on REDS4~\cite{REDS} in Sec~\ref{sec:qualitative_results}.

\section{Supplementary Implementation Details.}
\label{sec: implement_details}
For both recurrent-based and sliding-window-based methods, the number of frames used in each temporal stage is fixed regardless of the original video sequence.
Here, we take a temporal stage with a temporal size of $n$ as an example.
Given an input video sequence with 100 frames, we randomly sample a segment with a total frame of $n$.
Then the $n$ consecutive frames are fed into the VSR networks.
Note that the videos in  Vimeo-90K dataset~\cite{vimeo-90k} only contain 7 frames.
When the temporal size is larger than 7, we flip the videos and then concatenate them to the original videos.
Thus the total number of frames will be 14, which is the largest temporal size used in our temporal cycle.

\section{Additional Spatial-Temporal Stage}
\label{sec:addi_ST}
In our default setting, the spatial/temporal sizes in the last (largest) spatial-temporal stage are the same as baseline.
Here, we show that our multigrid training can achieve better performance with additional large spatial-temporal stages. Specifically, 
apart from the spatial/temporal sizes used in the main paper, we further train the VSR network~\cite{chan2021basicvsr} with an addition large spatial-temporal stage for 20,000 iterations, where the spatial and temporal sizes are set to {\color[HTML]{656565} 72$\times$72} and {\color[HTML]{656565} 17}, respectively. As shown in Table~\ref{tab:ab_st}, with the additional spatial-temporal stage ({\color[HTML]{656565} 72\&17}), multigrid training yields better performance (\ie, $+0.08$ dB) than baseline. However, the training is much slower than our default setting. Therefore, in order to achieve a better trade-off between performance and efficiency, we adopt the strategy in Table~\ref{tab:ab_st}\ding{173} as our default setting.

\begin{table}[h]
	\centering
	\tablestyle{4pt}{1.1}
	\begin{tabular}{l|x{32}cx{32}x{32}x{32}}
		model & index & spatial \& temporal size & iteration & speedup & PSNR  \\
		\shline
		baseline & \ding{172} &64\&15  & 300k & - & 30.91 \\
		\hline
		multigrid(\textbf{ours})  & \ding{173} & 32\&7 \textbf{/} 32\&11 \textbf{/} 32\&15 \textbf{/} 64\&7 \textbf{/} 64\&11 \textbf{/} 64\&15 &  75k & \textbf{6.2$\times$} & 30.90  \\
		multigrid & \ding{174} & 32\&7 \textbf{/} 32\&11 \textbf{/} 32\&15 \textbf{/} 64\&7 \textbf{/} 64\&11 \textbf{/} 64\&15 \textbf{/} {\color[HTML]{656565} 72\&17} & 95k &  3.4$\times$  & \textbf{30.98} \\
	\end{tabular}
	\vspace{7pt}
	\caption{{\bf Comparison of different combinations of spatial/temporal sizes.} All the results are obtained by BasicVSR-M on REDS4. The sizes are presented according to their orders in training. For example, `32\&7/32\&11' indicates that the training beings with a spatial size of $32\times32$ and a temporal size of $7$, and then is switched to a spatial size of $32\times 32$ and a temporal size of $11$. Our dynamic learning rate scheduler and large minibatch training are used in the multigrid variants. We report the wall-clock speedup relative to baseline.
		\label{tab:ab_st}
	}
\end{table}

\section{Qualitative Results}
\label{sec:qualitative_results}
In this section, we provide additional qualitative comparisons between our method and the baselines (BasicVSR~\cite{chan2021basicvsr}, BasicVSR-M, and EDVR-M~\cite{wang2019edvr}) on REDS4. As shown in Figure~\ref{fig:main_qualitative}, the methods with our multigrid training and large minibatch training obtain comparable visual results to the baselines. For example, both BasicVSR and BasicVSR (\textbf{ours}) successfully recover clear license plate number (first row in Figure~\ref{fig:main_qualitative}), and produce sharp edges (second and third rows in Figure~\ref{fig:main_qualitative}).

\begin{figure*}[h]
	\centering
	\includegraphics[width=\linewidth]{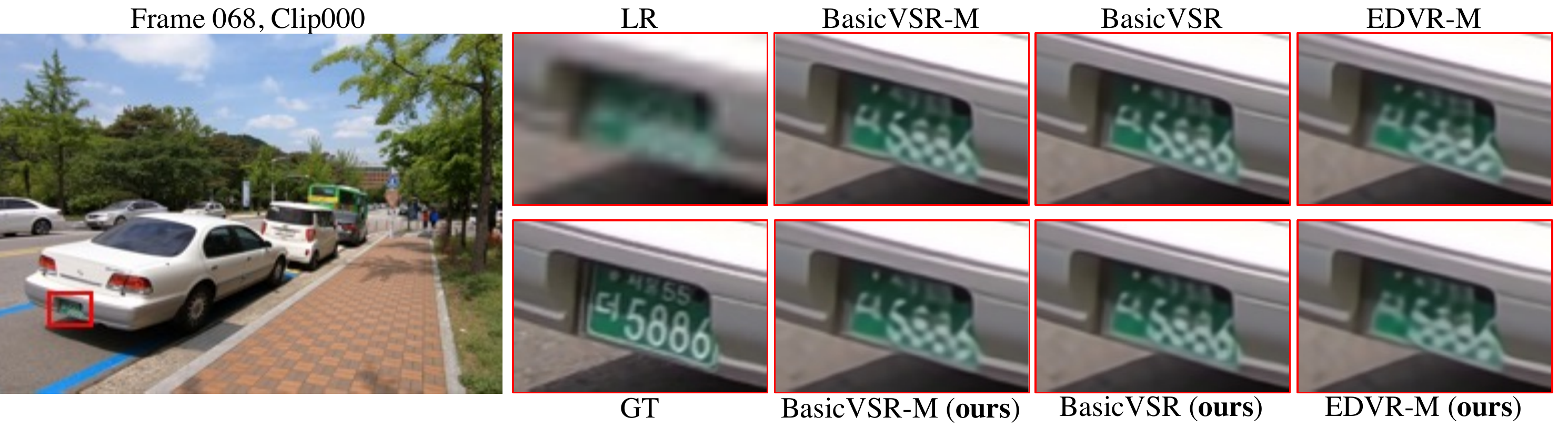}
	\includegraphics[width=\linewidth]{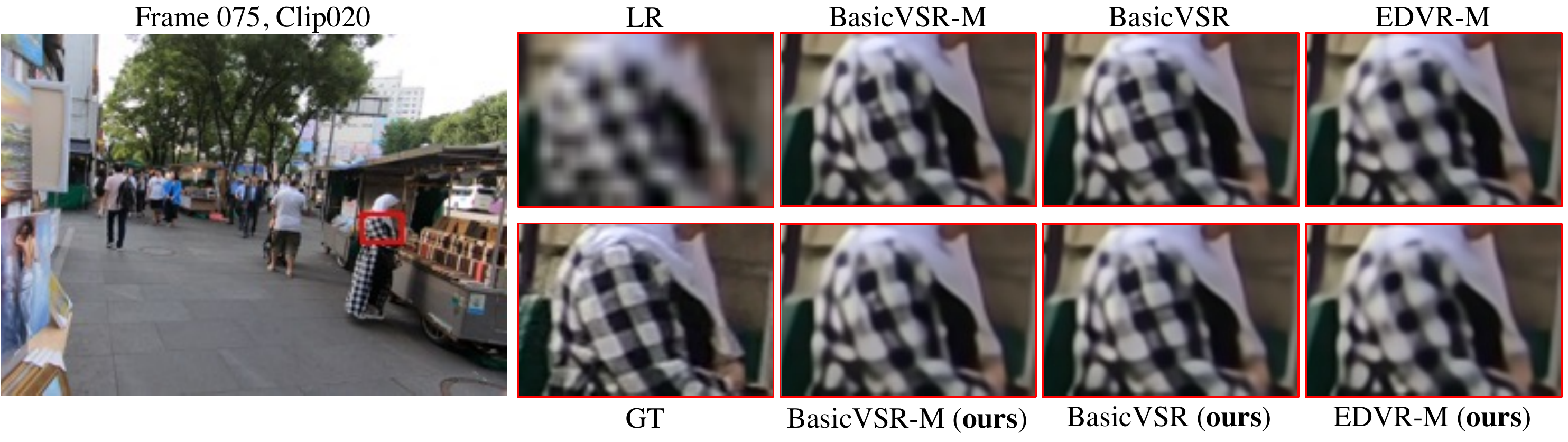}
	\includegraphics[width=\linewidth]{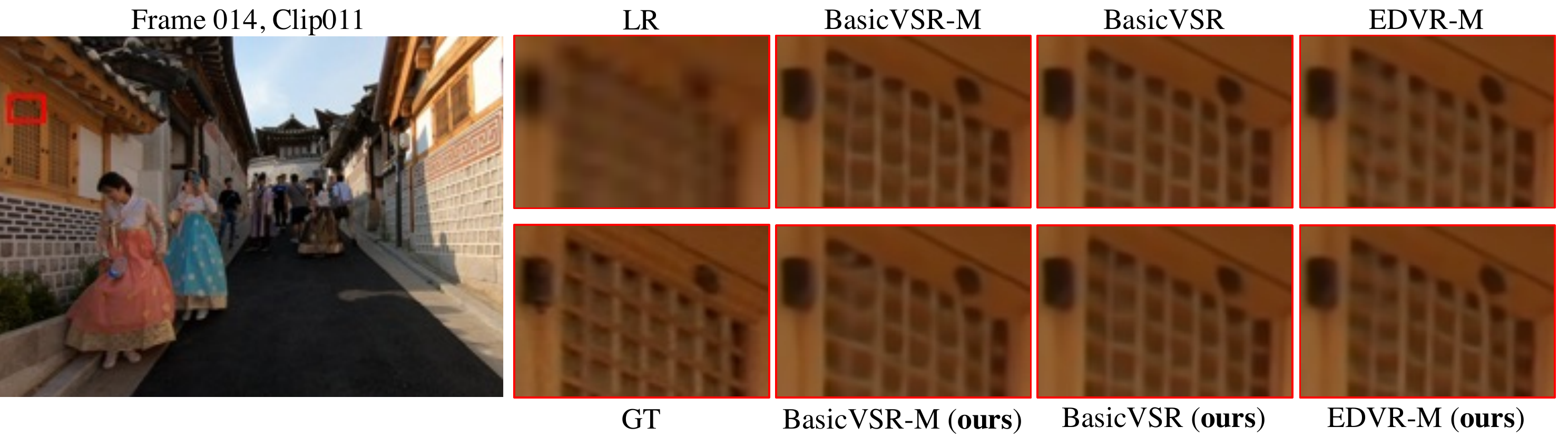}
	\vspace{5pt}
	\caption{{\bf Qualitative results on REDS4} for 4$\times$ VSR on BasicVSR-M, BasicVSR, and EDVR-M models. The methods with our proposed multigrid training and large minibatch training achieve comparable visual results to the baselines. \textbf{Zoom in for best view.} }
	\label{fig:main_qualitative}
\end{figure*}

\end{document}